# Evaluating Students' Open-ended Written Responses with LLMs: Using the RAG Framework for GPT-3.5, GPT-4, Claude-3, and Mistral-Large

**Jussi S. Jauhiainen [1] [2] and Agustín Garagorry Guerra [1]**

[1] Department of Geography and Geology, University of Turku
e-mail: jusaja@utu.fi
[2] Institute of Ecology and the Earth Sciences, University of Tartu

**ABSTRACT**

Evaluating open-ended written examination responses from students is an essential yet time-intensive task for educators, requiring a high degree of effort, consistency, and precision. Recent developments in Large Language Models (LLMs) present a promising opportunity to balance the need for thorough evaluation with efficient use of educators' time. In our study, we explore the effectiveness of LLMs—ChatGPT-3.5, ChatGPT-4, Claude-3, and Mistral-Large—in assessing university students' open-ended answers to questions made about reference material they have studied. Each model was instructed to evaluate 54 answers repeatedly under two conditions: 10 times (10-shot) with a temperature setting of 0.0 and 10 times with a temperature of 0.5, expecting a total of 1,080 evaluations per model and 4,320 evaluations across all models. The RAG (Retrieval Augmented Generation) framework was used as the framework to make the LLMs to process the evaluation of the answers. As of spring 2024, our analysis revealed notable variations in consistency and the grading outcomes provided by studied LLMs. There is a need to comprehend strengths and weaknesses of LLMs in educational settings for evaluating open-ended written responses. Further comparative research is essential to determine the accuracy and cost-effectiveness of using LLMs for educational assessments.

## 1. Introduction

Large Language Models (LLM) are being integrated into workflows by a variety of users, institutions, and stakeholders since the launch of ChatGPT-3.5 in November 2022. Subsequently, this and other LLMs have been adopted for many kinds of uses. The educational sector, in particular, has seen significant impact in this. Educators are using LLMs for tasks like adapting and creating content and evaluating students' performance, and students leverage these tools for assistance in writing essays and completing assignments. Despite their benefits, the adoption of LLMs also presents challenges. These include generating inaccurate ("hallucinated") content, security challenges with potential leakage of user data for model training and ethical issues for using the models inappropriately (Baidoo-Anu & Owusu Ansah 2023; Dai et al., 2023; Gimpel et al., 2023; Lo, 2023).

      A promising application within education is to use LLMs to assess students' open-ended responses. Traditionally, evaluating open-ended examination responses has been time-consuming and labor-intensive task for educators. However, if LLMs can effectively handle this task or at least support teachers in it, they could significantly reduce teacher workload, and potentially enhance their job

satisfaction and results. This could, in turn, improve the overall quality of the learning environment, benefiting both teachers and students. However, implementing LLMs in educational settings demands a thorough understanding of their capabilities and limitations to ensure that their integration helps to achieve educational goals while mitigating potential risks (Adiguzel et al., 2023; Bahroudn et al., 2023).

Research on the use of LLMs in education is still developing, with a focus primarily on commercially popular open-access models like ChatGPT-3.5. Nevertheless, there is a need to explore a wider array of LLMs. Our study addresses this gap by providing a comparative analysis of various LLMs' performance in educational contexts, focusing on integrating these tools intelligently, securely, transparently, and cost-efficiently. Additionally, our research introduces a methodological framework that could be adopted and refined in future studies to further enhance the effectiveness of LLMs in educational applications.

In this article, we explore the effectiveness of LLMs in assessing open-ended written examination responses from university students. We utilized the respective APIs for all the models included in this study: gpt-3.5-turbo, gpt-4-0125-preview, claude-3-opus-20240229, and Mistral-Large-large-latest. For ease of reference, these models will be referred in this article to as Gpt3.5, Gpt4, Claude3, and Mistral-Large-Large, respectively. Our investigation focuses on various factors that influence the feasibility of implementing these models in an educational setting, including the accuracy of their grading, consistency of grading results, processing speed, control over the model, and the costs associated with using computational resources.

Our investigation is structured around four primary research questions (RQs):

RQ1: What are the main characteristics of the evaluation process when open-ended written responses are evaluated and graded with Gpt3.5, Gpt4, Claude3, and Mistral-Large?

RQ2: What are the differences between the grades assigned by these LLMs?

RQ3: How consistent are the grades assigned by these LLMs?

RQ4: What were the processing times of these LLMs to perform the evaluation?

## 2. LLMs in Educational Environments

2.1. Elements behind LLMs' Capacity to Evaluate Written Texts

In general, LLMs are sophisticated deep learning algorithms and models excelling in tasks such as summarization, recognition, translation, prediction, and content generation, built on vast datasets for training (Wang et al., 2023). The educational sector has shown keen interest in LLM advancements, experiencing a blend of beneficial and challenging impacts (Baidoo-Anu & Owusu Ansah, 2023; Dai et al., 2023; Lo, 2023). This dual influence highlights the necessity for ongoing research and the development of robust frameworks to optimize the deployment of LLMs in educational settings for the benefit of both educators and learners. Three pivotal developments have catalyzed the rise of LLMs, particularly in the evaluation of written educational texts.

The first element supporting the LLM-based evaluation of written texts is the advancement in machine translation, where models utilize architectures that feature the ability to "soft-search" for relevant parts of a source sentence to predict the desired outcome. This breakthrough has significantly enhanced the model's ability to focus on relevant text segments, improving context processing capabilities (Bahdanau et al., 2014).

Another significant advancement is the development of the Transformer architecture by Google DeepMind (Vaswani et al., 2017). Transformers employ multi-head attention layers that enable the model to process various word characteristics simultaneously, thus enhancing both efficiency and performance. Unlike models based on recurrent or convolutional layers, Transformers can be trained more rapidly and are more scalable. LLMs based on this architecture use mechanisms like Temperature, Top-k and Top-p to generate diverse text outputs. The temperature parameter influences the randomness of predictions—lower values like 0.0 produce more predictable text, while higher values like 0.5 introduce greater creativity in text prediction. The Top-k sampling restricts the model to only consider the top 'k' probable next words, while the Top-p sampling uses a cumulative probability threshold to select the next words, adding flexibility and nuance to text generation (Cohere et al., 2022).

The third, and currently the most visible implementation of LLMs emerged with the launch of OpenAI's ChatGPT-3 in November 2022. This model combines the interactive format of chatbots with the generative capabilities of an LLM and the robust processing power of the Transformer architecture. This combination of interactivity, generative capacity, and sophisticated architecture enhances LLMs' utility in contemporary applications, notably in educational settings where they can be leveraged for tasks such as evaluating students' written texts, facilitating adaptive learning environments, and providing automated feedback (Baidoo-Anu & Owusu Ansah 2023; Dai et al., 2023; Jauhiainen & Garagorry Guerra, 2023; Lo, 2023; Wang et al., 2023). These capabilities underline the transformative potential of LLMs in education, necessitating careful consideration of their deployment to maximize benefits while addressing inherent challenges efficiently.

2.2. Assessing Open-ended Written Responses with LLMs

The assessment of open-ended responses for examination questions is one of the most labor-intensive aspects of educational evaluation. While these evaluations are necessary, they can be burdensome for educators, especially when applied to large groups of students. As in all evaluation activity, there are also risks of human errors and subjectivity as well as interpretation differences between human evaluators.

To alleviate these challenges, automated computer-assisted evaluation systems have been developed. Historically, the automation of open-ended answer evaluation has incorporated technologies such as Convolutional Neural Networks (CNNs), Long Short-Term Memory networks (LSTMs), and Transformers, including implementations such as using BERT for automated scoring (Miltsakaki & Kukich, 2004; Chen & He, 2013; Beseiso et al., 2020).

LLMs have proven versatile in various educational settings. For instance, in previous work, we utilized LLMs to create adaptive learning environments that dynamically adjusted learning materials to match the learner's skill level, thereby enhancing student engagement with the learning materials (Jauhiainen & Garagorry Guerra, 2023). In other forthcoming articles, we have addressed recalling, evaluation and feedback processes of LLMs.

The release of ChatGPT-3.5 sparked particular interest in its use for assessing students' written texts, such as open-ended exam responses—a task traditionally demanding significant teacher involvement. Most studies so far have focused on discussing the potential of generative AI technologies rather than analyzing systematically LLMs' concrete effects on educational practices (Baidoo-Anu & Owusu Ansah 2023; Dai et al., 2023; Gimpel et al., 2023). Recent studies have also explored ChatGPT-3.5's potential in providing articulate and high-quality responses to open-ended questions (Guerra et al., 2023; Vázquez-Cano et al., 2023), for providing detailed feedback for students (Bewersdorff et al., 2023) and supporting more generally higher education students (Bernabei et al., 2023), demonstrating how it has been used in wide educational contexts.

Employing LLMs like ChatGPT for evaluating students' written texts requires a systematic approach. LLMs need to access the students' examination materials and corresponding questions. They must accurately interpret students' responses and adhere to educational evaluation guidelines. Finally, they need to follow the grading system and assign grades based on correct and consistent grading.

However, the use of LLMs in educational settings is not without its challenges. Primary concerns regard security and ethics. The material inserted to LLM might be used for training that LLM if not stated otherwise. Another significant issue is "hallucinations," where LLMs generate factually inaccurate information or that its recalled texts contain something that was not initially there. As these inaccuracies are presented as valid, they can mislead both students and educators, posing a considerable risk in educational contexts. The presence of hallucinations needs to be systematically tested and eliminated (McIntosh et al. 2023). Furthermore, LLMs potentially inadvertently produce harmful or inappropriate content. To address this, recent models have incorporated safety reward signals during reinforcement learning with human feedback (RLHF) to reduce undesirable outputs (OpenAI, 2023). In addition, LLMs are typically trained offline, which means they do not learn from real-time interactions during deployment. This approach minimizes the risk of real-time errors but can lead to performance degradation over time as the model becomes outdated (Géron et al., 2022). A critical aspect of using LLMs involves addressing and mitigating biases that stem from their training data, architecture, and hyperparameter settings. Additionally, many ethical and secure everyday uses of LLMs remain unaddressed. Consequently, many educational institutions have expressed concerns about the use of LLMs, with some limiting or even prohibiting their use in education (Johnson, 2023).

## 3. Material and Methods

The novelty and rapid on-going development of LLMs suggest that extensive testing is still required regarding content types and use cases across different languages. This article demonstrates the utility of comparing various LLM tools to determine the most suitable options based on specific educational needs, particularly for evaluating and grading students' open-ended responses. Such comparative analysis is crucial to identify the best-fit models for particular educational applications.

The study was based on 54 open-ended responses from students enrolled in a master's level geography course at the University of Turku, taught in English. The professor, who is an expert in the subject matter, selected three scholarly articles to form the basis of the questions designed to assess the students' understanding about the learning material after they read it. Each article was accompanied by three corresponding questions about the content of the learning material. This Is a common practice at the university. This approach replicates a typical university exam setting where students are required to demonstrate their knowledge within a specified time. For the test, we used only text-based materials to eliminate any potential "noise" that visual elements like images or tables might introduce when using LLMs for evaluation as not all models perform enough well in multimodal contexts.

Our methodology for evaluating the performance of LLMs in assessing student responses involved several structured steps. Firstly, we obtained consent from all participating students to include their responses in the study. To ensure anonymity, the professor leading the study anonymized the data by removing all personal identifiers, thereby maintaining the anonymity of the subjects when analyzed with the LLMs. We did not collect any sensitive information such as gender, age, or country of origin, focusing solely on the capability of the LLMs to process and evaluate student responses accurately.

Secondly, we collected the student answers and utilized the Langchain Open AI Embedding method to convert the text into numerical representations. This was part of our data generation process. Reference texts were first processed using the PyPDF library, which allowed us to segment documents into 500-token chunks with a 20-token overlap. This specific granularity was selected to provide the LLMs with adequate context for accurately evaluating the student responses, while preventing token overflow in model inputs. For each student's response, we calculated cosine similarities with the document chunks, selecting the top five most relevant chunks (k=5). These chunks were then reorganized to optimize the retrieval process, thus enhancing the LLM's efficiency in referencing pertinent information during the evaluation process as outlined by Liu et al. (2023). We followed the RAG (Retrieval-Augmented Generation) technique (Figure 1). The use of RAG is useful in applications where the quality of output benefits from specific information, such as in question answering systems, content creation, and advanced chatbot functionalities. We mainly focused on Temperature and Prompt with RAG, which are among the most common approaches when implementing LLMs in development.

Thirdly, to ensure the reliability of our research prompts, we adhered to established standards such as verification-based chain-of-thought (CoT) prompting (Wei et al., 2022) along with other advanced prompting techniques. This rigorous approach helped us to confirm the consistency and validity of the prompts used for the LLMs' evaluation of the students' responses. Various prompts were tested before selecting the final ones used. In our methodology, each LLM was finally provided with the same prompts, reference materials, individual student responses, corresponding questions, and detailed evaluation guidelines. This comprehensive setup equipped the LLMs with the necessary context to accurately assess the students' knowledge (Figure 1).

Additionally, the prompts were customized to reflect the specific educational context, indicating roles such as the "University Professor" and specifying the academic level as master's degree. This customization was integral to tailoring the evaluation process to the unique requirements of the educational setting. The utilized evaluation criteria, validated by a diverse group of educators including teachers, pedagogical experts, and university staff, were aimed at assessing not only the overall grading accuracy of the LLMs but also specific aspects in the student written responses such as the completeness of answers, factual accuracy, logical consistency, context relevance and grammar and spelling. These parameters were essential for a detailed and comprehensive assessment of the LLMs' performance in educational applications (Figure 1).

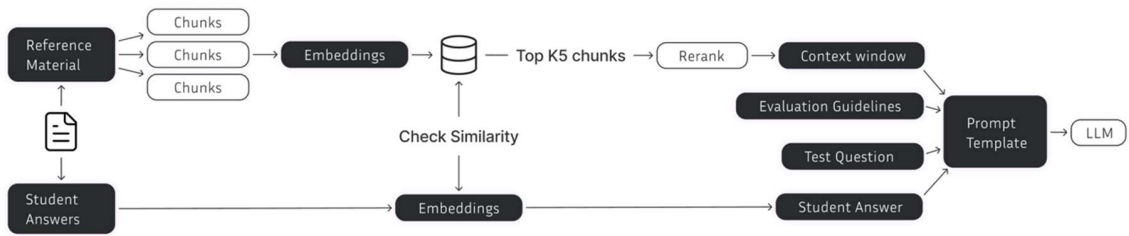

Figure 1. Evaluation process for generative AI LLM evaluation with reranked RAG.

Fourthly, we adopted a grading scale based on the Finnish higher education system, which includes grades Fail (0), Passable (1), Satisfactory (2), Good (3), Very Good (4), and Excellent (5). This scale was familiar to the participating university, teachers, and students, thereby enhancing the usability and relevance of our results. Each grade was initially presented in verbal form and subsequently converted into numerical values during the data cleaning phase for more straightforward analysis. This grading scale was applied to evaluate both the overall final grade of each answer and the specific evaluation parameters used by the LLMs to assess the answers.

Fifthly, we conducted multiple evaluations for each student's answer using a 10-shot scenario, where each LLM assessed each answer 10 times. In this context, the term "shot" refers to the repeated, independent processing of a student's answer by an LLM, which allowed us to observe and document variability in performance across multiple iterations. This method of repeating several times the evaluation was vital for identifying potential inconsistencies in the models, which could lead to biased or inaccurate evaluations if the evaluation or grading would have been inconsistent for any reason. This iterative evaluation process helped us to identify patterns and characteristics in each LLM's behavior. This would provide useful insights for educators, students, and researchers considering the use of these tools in educational settings. Among the data, there were 23 cases in which the LLM did not follow the prompt as expected, resulting in a lack of evaluations. These cases are distributed as follows: one partial case for Mistral-Large (specifically for individual parameters, the final grade was provided), 11 cases for GPT-3.5 at temperature 0.0, and 11 cases for GPT-3.5 at temperature 0.5. In both instances, the model failed to provide an evaluation. These cases represent 0.5% of the total evaluations, to which no significant impact is attributed.

Sixthly, to quantitatively assess and compare the performance differences among the LLMs, we employed Python programming language and Pandas library. Our statistical analysis primarily focused on descriptive statistics, which included calculating percentage distributions, means, and standard deviations to summarize the data effectively. Additionally, we conducted cross-tabulation analyses to delve deeper into the relationships between different variables. These cross-tabulations helped us determine the statistical significance of our findings, utilizing Pearson's chi-square test and Spearman's rank correlation. These statistical tools allowed us to evaluate and highlight the nuanced performance variations between the LLMs, providing a robust framework for understanding their capabilities and limitations in educational assessment contexts.

## 4. Results

As highlighted in the introduction, a primary objective of our analysis was to delineate the key features of the student open-ended answer evaluation process using Gpt3.5, Gpt4, Claude3, and Mistral-Large-Large. We specifically focused on discerning the differences in the grades assigned by these LLMs, examining the reliability and consistency of grading results, and assessing the processing time required by each LLM during the evaluation.

The test involved analyzing three open-ended answers for questions derived from one reference material, with a total of three reference materials and nine questions used for the test. The answers in English spanned a length from 24 to 256 words with average length of 152 words. Each LLM was tasked with evaluating each of the 54 different student answers 10 times under a 10-shot scenario at temperature settings of 0.0 and another 10 times at the temperature setting of 0.5. Consequently, each LLM conducted a total of 540 + 540 evaluations, cumulating in a grand total of 4,320 (of which 23 were truncated or incomplete) evaluations of student answers having in total 656,640 words.

4.1. Differences in Grading Across LLMs

Four LLMs used for this test detected variations in the quality of students' open-ended answers, resulting in the assignment of different final grades. Notably, when the models operated under temperature settings of 0.0 or 0.5, discrepancies in the grades assigned by the LLMs were evident (Table 1).

The most frequently assigned grade by all LLMs together was Satisfactory (2), attributed to 34.39% of the answers, followed by Passable (1) at 21.03%, Good (3) at 18.1%, and Very Good (4) at 11.05%. Grades at the extremes of the evaluation scale, such as Excellent (5) and Fail (0), were less common, given to 9.63% and 6.15% of student responses, respectively.

Using the temperature setting 0.0 tended to yield slightly more often grades at the both ends of the evaluation spectrum (either Fail or Excellent) compared to using the temperature 0.5. Furthermore, the proportion of answers rated as Good (3) was noticeably lower with temperature 0.0 than with temperature 0.5 (Table 1). With a higher temperature, LLMs can implement more creativity in their grading rather than following the prompt instructions rigorously. There were also significant differences in the grading outcomes across the different models, both overall and in their performance under the two temperature settings. These variations highlight the influence of model settings and selection on the assessment of student responses.

Table 1. Final Grade given to student answers by LLMs (Gpt3.5, Gpt4, Claude3, and Mistral-Large-Large) along 10-shot evaluation with temperature 0.0 and 0.5 (%, number of cases, in total 4,298 evaluations).

|              | Fail – 0   | Passable - 1 | Satisfactory – 2 | Good – 3   | Very Good – 4 | Excellent – 5 |
|--------------|------------|--------------|------------------|------------|---------------|---------------|
| 0.0 (2149)   | 6.75 (145) | 19.64 (422)  | 35.18 (756)      | 17.4 (374) | 11.40 (245)   | 9.63 (207)    |
| 0.5 (2149)   | 5.58 (120) | 22.43 (482)  | 33.60 (722)      | 18.8 (404) | 10.7 (230)    | 8.89 (207)    |
| Total (4298) | 6.17 (265) | 21.03 (904)  | 34.39 (1478)     | 18.1 (778) | 11.05 (475)   | 9.26 (398)    |

The evaluation results in terms of grades assigned to students' answers by the studied LLMs show significant disparities (Table 2). For instance, when assigning the grade Fail (0), Gpt3.5 at temperature 0.0 assigned this grade to 14.37% of student answers, whereas Claude3 at temperature 0.5 did so for only 1.48% of answers. Additionally, Gpt4 issued at least three times as many Fail (0) grades as Mistral-Large-Large, and this discrepancy widened to over seven times when comparing Gpt3.5 with Claude3. Meanwhile, Gpt4 at temperature 0.0 awarded the grade Satisfactory (2) to nearly half (49.26%) of the answers, in stark contrast to Gpt3.5 at the same temperature, which assigned it to just a tenth (10.02%) of the answers. Regarding the highest grade, Excellent (5), Gpt3.5 at temperature 0.0 awarded this grade to almost a fifth of answers (20.04%), while Claude3 awarded no Excellent grades (0.0%) at either temperature setting. Mistral-Large's performance fell between these extremes, demonstrating a more moderate grading pattern compared to the other LLMs (Table 2).

Table 2. Final Grade given to student answers by different LLMs (Gpt3.5, Gpt4, Claude3, and Mistral-Large) along 10-shot evaluation with temperature 0.0 and 0.5 (%, number of cases, in total 4,298 evaluations).

|  | Fail – 0 | Passable – 1 | Satisfactory – 2 | Good - 3 | Very Good - 4 | Excellent – 5 |
|---|---|---|---|---|---|---|
| Claude3, 0.0 (540) | 1.67 (9) | 10.37 (56) | 48.33 (261) | 18.15 (98) | 21.48 (116) | 0.00 (0) |
| Gpt3.5, 0.0 (529) | 14.37 (76) | 34.59 (183) | 8.51 (45) | 13.42 (71) | 9.07 (48) | 20.04 (106) |
| Gpt4, 0.0 (540) | 7.41 (40) | 11.85 (64) | 49.26 (266) | 21.48 (116) | 5.37 (29) | 4.63 (25) |
| Mistral-Large,0.0 (540) | 3.70 (20) | 22.04 (119) | 34.07 (184) | 16.48 (89) | 9.63 (52) | 14.07 (76) |
|  |  |  |  |  |  |  |
| Claude3, 0.5 (540) | 1.48 (8) | 13.52 (73) | 43.33 (234) | 22.59 (122) | 19.07 (103) | 0.00 (0) |
| Gpt3.5, 0.5 (529) | 11.91 (63) | 36.86 (195) | 10.02 (53) | 12.67 (67) | 10.96 (58) | 17.58 (93) |
| Gpt4, 0.5 (540) | 7.41 (40) | 15.37 (83) | 46.85 (253) | 21.11 (114) | 4.63 (25) | 4.63 (25) |
| Mistral-Large,0.5 (540) | 1.67 (9) | 24.26 (131) | 33.70 (182) | 18.70 (101) | 8.15 (44) | 13.52 (73) |

As indicated with the results above, selecting one or another LLMs for evaluating students' answers can lead to significant discrepancies in grading outcomes, highlighting the importance of the model selection. For example, using one particular LLM might result in approximately one out of every seven answers receiving a failing grade (Fail, 0), presenting a substantial challenge for students performing at this level. In contrast, choosing another LLM could result in hardly any student answers (less than 1.5%) receiving a failing grade. At the other end of the spectrum, employing one specific LLM for evaluation could lead to nearly a fifth of the answers receiving the highest grade (Excellent, 5), whereas using another model could result in no student answers receiving an Excellent grade.

Additionally, the choice of temperature setting for the evaluation is crucial. In some instances, minor or no differences were observed when alternating between temperature settings of 0.0 and 0.5. However, in other cases, the proportion of certain grades awarded nearly doubled when the temperature setting was changed from 0.0 to 0.5. These variations underline the need for careful consideration in selecting both the LLM and the temperature settings to ensure fairness and accuracy in grading student responses.

Table 2 provides a detailed performance analysis of each model, but below is a generalization of each model's performance:

*Claude3* grading tended to be homogeneous. Between 48.33% (at temperature 0.0) and 43.33% (at temperature 0.5) of its evaluations categorized as the lower grade of Satisfactory (2). After this, the model exhibited a further tendency towards assigning grades in the center, with 18.15% (at temperature 0.0) to 22.59% (at temperature 0.5) of the evaluations falling into the Good (3) category. Claude3 generally assigned lower-middle-range grades, and notably, it never assigned the grade Excellent (5).

*Gpt3.5* displayed a distinct grading pattern, significantly different from the other models. On the one hand, a considerable proportion of Gpt3.5's evaluations—ranging from 34.59% (at temperature 0.0) to 36.86% (at temperature 0.5)—fell into the very low category of Passable (1). On the other hand, its grading spanned all grades from Fail (0) to Excellent (5), with a substantial distribution across each category.

*Gpt4* consistently assigned from lower to mid-range grades, primarily between Satisfactory (2) and Good (3), with almost half of its grades being Satisfactory (2)—49.26% (at temperature 0.0) to 46.85% (at temperature 0.5). This indicates a grading pattern that was somewhat similar to that of Claude3.

*Mistral-Large* demonstrated a broader distribution of grades across the spectrum, but particularly assigning lower grades of Satisfactory (2) and Passable (1). Specifically, 34.07% and 33.70% of its grades fell into the Satisfactory (2) category across temperatures 0.0 and 0.5.

The variability observed in the grading of the same answers by different LLMs despite all having the same reference material and instructions illustrates the necessity for educational organizations to be aware of the different behavior of LLMs. It is recommended to conduct comprehensive evaluations of multiple LLMs to determine the most suitable model that aligns with their specific grading standards and requirements, or at least to be aware of such LLM behavior. Comparative analyses are vital to ensuring fairness and consistency in automated grading systems of open-ended answers with LLMs. It will support institutions in tailoring their technology usage to reliable educational outcomes. Additionally, fine-tuning the model's grading approach through precise prompting, calibrating the model's evaluation results with coefficients or adjusting computational algorithms may provide viable solutions if educational organizations are limited in their choice of models for evaluation.

4.2. Comparison of LLMs along their Scoring Criteria

Our research assesses the grading performance of various LLMs in comparison to one another. The potential influence of prompt engineering, which could affect model performance, was controlled in our study through strict adherence to advanced prompting techniques and consistent research standards. All models received identical prompts, reference materials, and student answers. The uniformity of prompt instructions and the context provided to the LLMs suggest that the observed variation in model performance is likely due to inherent characteristics and settings of each model rather than the prompts themselves, this also explains specific cases like Gpt3.5 not respecting the output format.

Setting fully objective and transparent grading is inherently challenging. Ultimately, one needs to decide what is the correct and fair grade given to each answer. Comparing LLMs with a single human evaluator do not necessarily provide a solution for this because also human evaluators can make mistakes or could have tendency to assign lower or higher grades than the answers would merit.

In our study, we focused on LLMs and did not involve human evaluation into the process. Instead, we implemented a process in which all LLMs used took part in deciding what would be the proper grade for each answer. In this process, each answer received 40 evaluations (10 evaluations per model), and we selected as the reference standard grade the most frequently occurring grade (mode value) for each answer. This process was conducted separately for both temperature settings of 0.0 and 0.5.

With this "accurate" reference grade established as a benchmark, we analyzed the performance of each LLM against this standard using three key metrics: Accurate (where the LLM-generated grade perfectly matches the benchmark grade that was the most commonly given grade by all LLMs studied), Small Deviation (grades given by LLMs that were within ±1 of the benchmark grade), and Inaccurate (grades given by LLMs that were more distant than 1 grade from the benchmark grade). Grades deemed Accurate or within a Small Deviation range can be considered acceptable, reflecting the natural variability that might occur also among human evaluators.

However, we also found that in a few (13.88%) cases of evaluating the student's answers, the LLMs divided in indicating whether one grade or the grade next to it should be the correct one. These were the cases in which the most commonly suggested grade got at least 40% of all grades suggested by all LLMs and the second most common grade got at least 30% of all grades suggested by all LLMs. The share of such "undecisive" grades was rather low being 9.3% at temperature 0.0 (five cases). At temperature 0.5, they were 18.5% of cases (ten cases).

The primary aim of our analysis is to identify LLMs that not only maximize accuracy their grading but also minimize the incidence of significant grading discrepancies, thereby enhancing the reliability and fairness of automated grading systems with LLMs. The results are synthesized in Table 3.

*Claude3* at temperature 0.0 demonstrated a good alignment with the LLM benchmark grade with 62.78% of its grades falling into the Accurate category and a very high 87.04% share was within Small Deviation (±1 of the benchmark). However, at temperature 0.5, Claude3's performance became more distant from the benchmark grade with the share of accurate grades lowering to 48.70%, though those within ±1 of the benchmark was at 88.52%. Overall, Claude3's grading performance at temperature 0.0 can be considered good as barely-12.96% of evaluations fell more than one point out of the benchmark grade.

*Mistral-Large* (at a temperature of 0.0) performed well, achieving an accuracy of 56.67% of its grades being the same as the benchmark grade. Of its grades, 84.08% were within ±1 of the benchmark (Small Deviation). However, at temperature 0.5, Mistral-Large's performance became less accurate with the share of accurate grades dropping to 43.15% and those within ±1 of the benchmark at 81.11%. In both scenarios the share of 'Inaccurate cases' and 'Inaccurate plus bigger difference' increased when a temperature of 0.5 was assigned to 15.92% at temperature 0.0 and 18.89% at temperature 0.5.

*Gpt4* at temperature 0.0 had the second highest performance with 58.15% of its grades falling into the Accurate category and of its grades, 89.27% were within Small Deviation (±1 of the benchmark). At temperature 0.5, it became slightly more aligned toward the benchmark grade. The share of accurate grades increased to 59.07%, and slightly decreased within ±1 of the benchmark to 89.07%. Overall, Gpt4's grading performance can be considered particular: it managed to grade slightly less often fully accurate grades, especially at 0.0 temperature level, but almost nine out of ten of its grades were within small deviation from the benchmark grade.

*Gpt3.5* showed substantially weaker grading accuracy compared to other LLMs studied in this article. At temperature 0.5, less than a third (29.68%) of its grades fell into the Accurate category and within the category Small Deviation (±1 of the benchmark) were 64.65% of its grades. The performance of Gpt3.5 at temperature 0.0 did not improve, with only a quarter (24.95%) of its grades falling into the Accurate category and of its grades, within small deviation (±1 of the benchmark) were only 59.62%. Comparing the grading by Gpt3.5 to other LLMs studied, its performance was substantially inferior. Additionally, 13.04% evaluations were "wildly" different, deviating from the benchmark grade by more than 2 points, which at a 6-point grading scale is significant. In this test, Gpt3.5 was not found to be a reliable tool for grading student's open-ended responses.

Overall, of the studied LLMs, the share of grades considered inaccurate, i.e. deviating more than 1 grade from the benchmark grade at 0.0 temperature was 12.96% for Claude-3, 15.92% for Mistral-Large, 10.75% for Gpt4, and 39.88% for Gpt3.5.

Table 3. Score differences from the benchmark LLM grade for Final Grade of student answers, 10-shot evaluation with temperature 0.0 and 0.5 variants (%, number of cases, in total 4,298 evaluations; in green the highest performance results, in red the lowest performance results).

|  | Inaccurate | Minor Deviation | Accurate | Minor Deviation | Inaccurate | Inaccurate |
| --- | --- | --- | --- | --- | --- | --- |
|  | +2 | +1 | 0 | -1 | -2 | Other |
| Claude3, 0.0 (540) | 3.89 (21) | 13.89 (75) | 62.78 (339) | 10.37 (56) | 2.22 (12) | 6.85 (37) |
| Gpt3.5, 0.0 (529) | 11.91 (63) | 11.34 (60) | 24.95 (132) | 23.33 (126) | 14.93 (79) | 13.04 (69) |
| Gpt4, 0.0 (540) | 1.67 (9) | 15.19 (82) | 58.15 (314) | 15.93 (86) | 3.89 (21) | 5.19 (28) |
| Mistral-Large, 0.0 (540) | 4.07 (22) | 10.00 (54) | 56.67 (306) | 17.41 (94) | 2.22 (12) | 9.63 (52) |
| Claude3, 0.5 (540) | 6.30 (34) | 25.19 (136) | 48.70 (263) | 14.63 (79) | 1.67 (9) | 3.52 (19) |
| Gpt3.5, 0.5 (529) | 13.23 (70) | 10.96 (58) | 29.68 (157) | 24.01 (127) | 13.23 (70) | 8.88 (47) |
| Gpt4, 0.5 (540) | 3.89 (21) | 13.52 (73) | 59.07 (319) | 16.48 (89) | 2.22 (12) | 4.81 (26) |
| Mistral-Large, 0.5 (540) | 8.89 (48) | 19.26 (104) | 43.15 (233) | 18.70 (101) | 4.26 (23) | 5.74 (31) |

### 4.3. Consistency of LLMs on Evaluating Student Open-ended Responses

Consistency is a critical attribute in the evaluation performance of LLMs for several reasons. Primarily, a high variance in the evaluation results of LLMs, such as the grades assigned, undermines the models' reliability for being consistent in their grading of students' open-ended responses and assigning final grades to their performance. To ascertain a model's consistency, it is necessary to run multiple evaluations of the same answer. In this study, we opted for a 10-shot scenario, evaluating each answer 10 times.

To verify the grading consistency of LLMs, we conducted a thorough analysis of 54 different student answers and examined the corresponding LLM evaluation results. Our approach involved analyzing whether the grades assigned by each LLM were consistent across all 10 shots, both for the final grade and for each parameter used in the evaluation. Consistency was defined as having identical grades within a 10-shot series for a particular student response analyzed at both 0.0 and 0.5 temperature settings. If all grades within a series were the same then, the LLM demonstrated full consistency for that answer. Conversely, any variation within the series indicated a lack of full consistency in grading by the LLM. This method allowed us to systematically determine the reliability of each LLM in maintaining grading standards across multiple evaluations.

Mistral-Large at 0.0 temperature demonstrated the highest grading consistency among the models, with 83.33% (45 out of 54) of its 10-shot gradings showing no variation within the evaluations. Claude3 showed considerable consistency as well, with 70.37% (38 out of 54) of its 10-shot gradings displaying no internal variation. Gpt4 and Gpt3.5 had lower consistency rates, with 35.19% (19 out of 54) and 18.52% (10 out of 54) respectively, showing uniformity in the grades assigned.

When the temperature setting was increased to 0.5, the consistency of the LLMs noticeably decreased, highlighting a sensitivity to temperature changes. At this higher temperature, Gpt4 achieved the highest level of consistency, albeit reduced to 20.37% (11 out of 54). Mistral-Large's consistency significantly declined to 14.81% (8 out of 54). Both Claude3 and Gpt3.5 demonstrated low consistency, with 12.96% (7 out of 54) and 14.81 (8 out of 54) of their gradings being internally consistent, indicating substantial internal variation in more than 85% of their evaluations.

Overall, across all evaluation parameters, Mistral-Large at 0.0 temperature consistently demonstrated high performance, with consistency percentages ranging from 87.04% to 88.89%. Claude3 followed, with consistency varying from 61.11% to 75.93%. Gpt4 showed lower consistency, ranging from 24.07% to 53.70%, while Gpt3.5 had the least consistency, varying between 16.67% and 37.04%. The increase in temperature to 0.5 led to reduced consistency across all models, with Mistral-Large remaining the most consistent but experiencing a significant drop, showing consistency percentages now between 24.07% and 33.33%. Gpt3.5 had the lowest consistency, ranging from 7.41% to 12.96%.

In terms of grading variability at 0.0 temperature, Mistral-Large and Claude3 displayed high consistency between 98% and 100% of consistency of its gradings within one grade point. Gpt4's consistency in final grades was also high when including one grade deviations reaching 90.75% consistency, yet it included notable outliers where the grade assignment was up to 4 points difference. Gpt3.5 demonstrated significant inconsistency with almost half (46.3%) of its final grade evaluations within a one-point deviation.

With temperature 0.5, grading variability increased across all models. In final grades, Claude3 maintained the highest consistency with 83.33% of its gradings falling within one grade point, while Gpt3.5 exhibited significant inconsistency with only 33.33% of its gradings within the same range, pointing to substantial fluctuations in its grading of student answers across all parameters.

Executing each test consumes significant computational resources, energy, and time, and utilizes tokens, which has economic implications. Therefore, model consistency is crucial. A more consistent model requires fewer iterations for reliable evaluations, reducing the computational power and time needed for each assessment. However, to establish a model's consistency, multiple trials are necessary. Fewer trials indicate higher consistency, but relying on a single trial could lead to arbitrary and potentially incorrect results, highlighting the balance between resource usage, grading reliability and type of task (summarize, evaluate, modify content, etc.).

In our study, which evaluated student answers both in terms of final grades and several related parameters, Mistral-Large at 0.0 temperature showed the highest consistency in grading. It consistently assigned the same grade in 83.33% of all 10-shot evaluations with minimal variation. The next most consistent model at 0.0 temperature was Claude3, which maintained a consistency level of 70.37%. Other models, such as Gpt4 and Gpt3.5, showed greater inconsistency, particularly at the higher temperature setting of 0.5, where their consistency in final grades dropped significantly to 20.37% and 12.96% respectively.

Looking at the evaluation of various parameters that assess different aspects of student answers—including context relevance, factual accuracy, completeness, logical consistency, and grammar and spelling—the patterns of consistency largely mirrored those observed in the final grade assessments. At 0.0 temperature, Mistral-Large exhibited very high consistency across all parameters, with rates ranging from 87.04% to 88.89%. Claude3 at 0.0 temperature was the second most consistent, with rates between 68.52% and 75.93%. In contrast, at 0.5 temperature, Gpt3.5 demonstrated notably lower consistency, ranging from 7.41% to 12.96% across the parameters. Gpt4 at 0.5 temperature showed higher variability in consistency rates, from 7.41% to 48.15%, though it performed notably better in the specific evaluation of grammar and spelling, reaching a consistency rate of 48.15%, although still lower compared to Claude3 and Mistral-Large.

Table 4. Share of cases without variation within their 10 shot-grading regarding the final grade and evaluation parameters (%, number of cases, in total 54 evaluation groups.

|  | **Range** | Context Relevance | Factual Accuracy | Completeness | Logical Consistency | Grammar & Spelling | Final Grade |
|---|---|---|---|---|---|---|---|
| Claude3, 0.0 | 0 | 74.93 (41) | 75.93 (41) | 74.07 (40) | 68.52 (37) | 61.11 (33) | 70.37 (38) |
|  | 1 | 24.07 (13) | 22.22 (12) | 24.07 (13) | 31.48 (17) | 33.33 (18) | 27.78 (15) |
|  | 2 | 0.00 (0) | 1.85 (1) | 1.85 (1) | 0.00 (0) | 3.70 (2) | 1.85 (1) |
|  | 3 | 0.00 (0) | 0.00 (0) | 0.00 (0) | 0.00 (0) | 1.85 (1) | 0.00 (0) |
|  | 4 | 0.00 (0) | 0.00 (0) | 0.00 (0) | 0.00 (0) | 0.00 (0) | 0.00 (0) |
|  | 5 | 0.00 (0) | 0.00 (0) | 0.00 (0) | 0.00 (0) | 0.00 (0) | 0.00 (0) |
| Gpt3.5, 0.0 | 0 | 31.48 (17) | 20.37 (11) | 37.04 (20) | 20.37 (9) | 16.67 (9) | 18.52 (10) |
|  | 1 | 40.74 (22) | 44.44 (24) | 33.33 (18) | 44.44 (24) | 44.44 (24) | 37.04 (20) |
|  | 2 | 24.07 (13) | 27.78 (15) | 20.37 (11) | 22.22 (17) | 31.48 (17) | 31.48 (17) |
|  | 3 | 3.70 (2) | 5.56 (3) | 5.56 (3) | 7.41 (3) | 5.56 (3) | 5.56 (3) |
|  | 4 | 0.00 (0) | 1.85 (1) | 1.85 (1) | 9.26 (1) | 1.85 (1) | 7.41 (4) |
|  | 5 | 0.00 (0) | 0.00 (0) | 1.85 (1) | 0.00 (0) | 0.00 (0) | 0.00 (0) |
| Gpt4, 0.0 | 0 | 37.04 (20) | 24.07 (13) | 37.04 (20) | 24.07 (13) | 53.70 (29) | 35.19 (19) |
|  | 1 | 48.15 (26) | 64.81 (35) | 46.30 (25) | 64.81 (35) | 24.07 (13) | 55.56 (30) |
|  | 2 | 9.26 (5) | 7.41 (4) | 14.81 (8) | 9.26 (5) | 16.67 (9) | 5.56 (3) |
|  | 3 | 5.56 (3) | 3.70 (2) | 1.85 (1) | 1.85 (1) | 3.70 (2) | 0.00 (0) |
|  | 4 | 0.00 (0) | 0.00 (0) | 0.00 (0) | 0.00 (0) | 1.85 (1) | 3.70 (2) |
|  | 5 | 0.00 (0) | 0.00 (0) | 0.00 (0) | 0.00 (0) | 0.00 (0) | 0.00 (0) |
| Mistral-Large, 0.0 | 0 | 88.89 (48) | 88.89 (48) | 87.04 (47) | 88.89 (48) | 87.04 (47) | 83.33 (45) |
|  | 1 | 9.26 (5) | 9.26 (5) | 12.96 (7) | 9.26 (5) | 11.11 (6) | 16.67 (9) |
|  | 2 | 1.85 (1) | 1.85 (1) | 0.00 (0) | 1.85 (1) | 1.85 (1) | 0.00 (0) |
|  | 3 | 0.00 (0) | 0.00 (0) | 0.00 (0) | 0.00 (0) | 0.00 (0) | 0.00 (0) |
|  | 4 | 0.00 (0) | 0.00 (0) | 0.00 (0) | 0.00 (0) | 0.00 (0) | 0.00 (0) |
|  | 5 | 0.00 (0) | 0.00 (0) | 0.00 (0) | 0.00 (0) | 0.00 (0) | 0.00 (0) |
| Claude3, 0.5 | 0 | 33.33 (18) | 22.22 (12) | 33.33 (18) | 33.33 (18) | 31.48 (17) | 12.96 (7) |
|  | 1 | 59.26 (32) | 53.70 (29) | 53.70 (29) | 50.00 (27) | 37.04 (20) | 70.37 (38) |
|  | 2 | 7.41 (4) | 22.22 (12) | 12.96 (7) | 16.67 (9) | 24.07 (13) | 16.67 (9) |
|  | 3 | 0.00 (0) | 1.85 (1) | 0.00 (0) | 0.00 (0) | 5.56 (3) | 0.00 (0) |
|  | 4 | 0.00 (0) | 0.00 (0) | 0.00 (0) | 0.00 (0) | 1.85 (1) | 0.00 (0) |
|  | 5 | 0.00 (0) | 0.00 (0) | 0.00 (0) | 0.00 (0) | 0.00 (0) | 0.00 (0) |
| Gpt3.5, 0.5 | 0 | 11.11 (6) | 9.26 (5) | 12.96 (7) | 9.26 (5) | 7.41 (4) | 12.96 (7) |
|  | 1 | 16.67 (9) | 24.07 (13) | 44.44 (24) | 24.07 (13) | 20.37 (11) | 20.37 (11) |
|  | 2 | 35.19 (19) | 37.04 (20) | 18.52 (10) | 38.89 (21) | 40.74 (22) | 33.33 (18) |
|  | 3 | 35.19 (19) | 25.93 (14) | 16.67 (9) | 22.22 (12) | 29.63 (16) | 20.37 (11) |
|  | 4 | 1.85 (1) | 3.70 (2) | 7.41 (4) | 5.56 (3) | 0.00 (0) | 12.96 (7) |
|  | 5 | 0.00 (0) | 0.00 (0) | 0.00 (0) | 0.00 (0) | 1.85 (1) | 0.00 (0) |
| Gpt4, 0.5 | 0 | 18.52 (10) | 7.41 (4) | 16.67 (9) | 22.22 (12) | 48.15 (26) | 20.37 (11) |
|  | 1 | 53.70 (29) | 68.52 (37) | 59.26 (32) | 44.44 (24) | 22.22 (12) | 59.26 (32) |
|  | 2 | 22.22 (12) | 22.22 (12) | 12.96 (7) | 29.63 (16) | 18.52 (10) | 16.67 (9) |
|  | 3 | 3.70 (2) | 0.00 (0) | 9.26 (5) | 3.70 (2) | 9.26 (5) | 3.70 (2) |
|  | 4 | 1.85 (1) | 1.85 (1) | 1.85 (1) | 0.00 (0) | 1.85 (1) | 0.00 (0) |
|  | 5 | 0.00 (0) | 0.00 (0) | 0.00 (0) | 0.00 (0) | 0.00 (0) | 0.00 (0) |
| Mistral-Large, 0.5 | 0 | 33.33 (18) | 29.63 (16) | 35.19 (19) | 24.07 (13) | 25.93 (14) | 14.81 (8) |
|  | 1 | 37.04 (20) | 38.89 (21) | 40.74 (22) | 46.30 (25) | 44.44 (24) | 42.59 (23) |
|  | 2 | 27.78 (15) | 31.38 (17) | 20.37 (11) | 27.78 (15) | 27.78 (15) | 35.19 (19) |
|  | 3 | 1.85 (1) | 0.00 (0) | 3.70 (2) | 1.85 (1) | 0.00 (0) | 7.41 (4) |
|  | 4 | 0.00 (0) | 0.00 (0) | 0.00 (0) | 0.00 (0) | 1.85 (1) | 0.00 (0) |
|  | 5 | 0.00 (0) | 0.00 (0) | 0.00 (0) | 0.00 (0) | 0.00 (0) | 0.00 (0) |

Table 5. Evaluation cases without any variation within their 10-shot evaluation.

| | Context Relevance | Factual Accuracy | Completeness | Logical Consistency | Grammar & Spelling | Final Grade |
|---|---|---|---|---|---|---|
| Claude3, 0.0 | 74.93 | 75.93 | 74.07 | 68.52 | 61.11 | 70.37 |
| Gpt3.5, 0.0 | 31.48 | 20.37 | 37.04 | 20.37 | 16.67 | 18.52 |
| Gpt4, 0.0 | 37.04 | 24.07 | 37.04 | 24.07 | 53.70 | 35.19 |
| Mistral-Large, 0.0 | 88.89 | 88.89 | 87.04 | 88.89 | 87.04 | 83.33 |
| Claude3, 0.5 | 33.33 | 22.22 | 33.33 | 33.33 | 31.48 | 12.96 |
| Gpt3.5, 0.5 | 11.11 | 9.26 | 12.96 | 9.26 | 7.41 | 12.96 |
| Gpt4, 0.5 | 18.52 | 7.41 | 16.67 | 22.22 | 48.15 | 20.37 |
| Mistral-Large, 0.5 | 33.33 | 29.63 | 35.19 | 24.07 | 25.93 | 14.81 |

The results highlight a significant influence of temperature settings on improving consistency, although this did not uniformly affect all models. Models like Mistral-Large and Claude3 showed substantial sensitivity to temperature adjustments. Gpt4 did not parallel this pattern, displaying almost a 50% difference in performance metrics. This disparity illustrates the nuanced behaviors of different LLMs under varying operational settings and emphasizes the importance of selecting the right model and settings for specific assessment needs.

4.4. Differences in Processing Time in Evaluation Tasks among Different LLMs

Processing time is an essential consideration when deploying LLMs in educational settings, particularly for the efficient and timely evaluation of large volumes of student assessments. In our analysis of LLMs' ability to grade open-ended questions, we define processing time as the duration each model requires to analyze context information and evaluate student responses thoroughly. This aspect is crucial for minimizing errors such as hallucinations and ensuring well-informed grading decisions.

To maintain the integrity of our data, we excluded substantial outliers from our analysis. These outliers represented cases of which evaluation and grading that required exceptionally long processing times—more than 10 times the average or exceeding 150 seconds, with one instance surpassing 240 seconds. While these cases, very few in total, were not considered in the primary analysis, their potential impact on computational costs could be significant, especially when considering the large-scale deployment of LLMs in academic assessments.

Our study found considerable differences in processing times both between different LLMs and between temperature settings within the same model. We observed that increasing the temperature setting from 0.0 to 0.5 typically resulted in a 4% to 12% increase in processing time for depending on the model. However, the correlation between processing times and the grade performance assigned by LLMs was relatively low, indicating that quicker processing does not necessarily translate to more accurate or consistent grading (Table 6).

The fastest average processing time per answer was recorded at 3.52 seconds by Gpt3.5 at temperature 0.5, followed closely by the same model at temperature 0.0 (3.89 seconds). In stark contrast, the slowest average processing times were significantly longer, with Claude3 at temperature 0.5 taking 21.55 seconds and Gpt4 at temperature 0.5 taking 20.72 seconds. This discrepancy implies that computational costs and associated energy consumption can vary more five times between the fastest and slowest models (Table 6).

To put this into perspective, evaluating 100 student answers with a single shot at time from the fastest model would take approximately 6.5 minutes, and using a 10-shot scenario would extend this time to about 1 hour and 5 minutes. Conversely, using the slowest model for a one-shot evaluation of the same number of answers would take nearly 36 minutes, and a 10-shot scenario would require almost 6 hours, if shots would be analyzed one after another. However, there are techniques to analyze several shots simultaneously, so these total processing times are only examples of extremely rare circumstances.

The variability in processing times was the largest in Gpt4 (both temperatures) and the smallest in Mistral-Large (both temperatures), as illustrated by the standard deviations. This signifies the necessity of a balanced approach in selecting LLMs for educational use. While faster processing times can improve the scalability of LLM applications, they must not compromise the accuracy and consistency of the outcomes. Therefore, educational institutions need to weigh both the efficiency and reliability of LLMs when integrating these technologies into their grading systems.

Table 6. Processing time of LLMs for evaluating student answers by one shot after another.

|  | mean | std | min | 25% | 50% | 75% | max | Pearson | Spearman |
|---|---|---|---|---|---|---|---|---|---|
| Claude3–0.0 | 20.61 | 3.82 | 12.70 | 18.33 | 20.12 | 22.12 | 66.25 | 0.028 | 0.034 |
| Gpt3.5 – 0.0 | 3.89 | 4.79 | 1.50 | 2.94 | 3.51 | 4.12 | 112.20 | -0.041 | -0.006 |
| Gpt4 – 0.0 | 18.51 | 9.28 | 7.80 | 13.54 | 16.78 | 21.46 | 120.09 | 0.009 | 0.096 |
| Mistral-Large – 0.0 | 11.17 | 2.91 | 5.92 | 9.17 | 10.64 | 12.40 | 27.57 | 0.177 | 0.145 |
| Claude3–0.5 | 21.55 | 4.09 | 14.44 | 18.81 | 20.89 | 23.54 | 48.55 | -0.007 | -0.027 |
| Gpt3.5 – 0.5 | 3.52 | 3.38 | 1.42 | 2.68 | 3.19 | 3.74 | 78.38 | 0.093 | 0.070 |
| Gpt4 – 0.5 | 20.72 | 6.74 | 9.25 | 15.84 | 19.74 | 24.21 | 64.30 | 0.132 | 0.135 |
| Mistral-Large – 0.5 | 11.95 | 3.05 | 6.16 | 9.87 | 11.36 | 13.80 | 34.08 | 0.247 | 0.255 |

5. Conclusions

This study demonstrated that LLMs like GPT-3.5, GPT-4, Claude-3, and Mistral-Large can be used to evaluate and grade students' open-ended written answers, especially when the Retrieval Augmented Generation (RAG) framework is utilized for the evaluation process. For a more consolidated grading, each answer has to be evaluated several times. The choice of temperature setting (0.0 or 0.5) significantly affects the performance of LLMs in evaluating open-ended responses, impacting both speed and consistency of grading; and the temperature setting 0.0 gave more consistent results. While GPT-3.5 processed responses the quickest, it showed less consistency in grading compared to other LLMs, and the range of grades it assigned was substantially broader. Comparative analyses between different LLMs will determine the most trustful and efficient LLMs that are suitable for evaluating students' open-ended responses.

The RAG framework is useful to implement efficient, ethical, and secure evaluation and grading of students' open-ended written responses with LLMs such as Gpt3.5, Gpt4, Claude3, and Mistral-Large. The Langchain Open AI Embedding method can be used to convert the text into numerical representations. Instead of a single evaluation of a student's written response, it is crucial that each answer is evaluated multiple times. We suggest 10 shots as the number of evaluations needed for each response to confirm the consistency of LLMs in grading.

The analysis revealed that different LLMs exhibited variability between the models in their grading results they assigned for students' open-ended responses. This reflects distinct training, architecture, and hyperparameters that these LLMs possess. It is crucial to be carefully aware which LLM will be used for evaluating students' open-ended answers, particularly given the rapid evolution of these models. It is advisable to conduct performance comparisons of various LLMs before their implementation for evaluation tasks. Additionally, the choice of the temperature setting (0.0 or 0.5) significantly influences the evaluation performance. It is recommended to use temperature 0.0 to minimize variability in grading results.

To test the grading performance of LLMs, it is critical to establish a reliable benchmark grade. There are fundamentally two methods to determine this. The first method involves employing multiple human experts to evaluate the answers and reach a consensus on the correct grade, typically using the mode value of their assessments. Relying solely on a single teacher or expert is not recommended, as grading is inherently subjective and may vary significantly between individuals. The second method is to use several high-performing LLMs to establish the benchmark grade, again favoring the mode value from multiple evaluations (such as a ten-shot scenario) to determine the most agreed-upon grading outcome. All participating LLMs need to be of sufficiently high quality to ensure reliable evaluation and grading results. Additionally, using the temperature setting 0.0 is advised to minimize the grading variability and enhance the consistency of the LLM-based evaluations.

In this test, we set as the benchmark grade that grade which the studied four LLMs most often selected in their evaluation. The evaluation grade that met exactly the benchmark grade was considered accurate, the grades that deviated 1 grade from the benchmark grade in a six-scale grading system were considered having small deviation but still reliable. The grades that were 2 grades from the benchmark grade were considered inaccurate. At 0.0 temperature, the share of inaccurate grades was 12.96% for Claude-3, 15.92 % for Mistral-Large, 10.75% for Gpt4, and 39.88% for Gpt3.5. However, Claude-3 was by far the most accurate when considering the share of grade that were same with the benchmark grade (62.78%), followed by Gpt-4 with 58.15%, 56.67% for Mistral-Large, and lastly 24.44% for Gpt3.5.

The LLMs showed differences in grading consistency and processing time. Longer processing times entail higher computational costs, making the evaluation process more expensive. Although the freely accessible GPT3.5 processed responses the fastest, it exhibited substantially less consistency in grading results compared to others, and assigned a broader range of grades, including many of those considered inaccurate. Consequently, the use of Gpt3.5 for systematic evaluation of students' open-ended answers is not recommended as it has substantial inconsistency in its grading.

LLMs are poised to become integral tools in educational settings, aiding in the evaluation of student performance, including open-ended responses. By early 2024, criticism from scholars and educators so far has primarily focused on LLMs seemingly inconsistent and unsecure performance in evaluation. However, most cases documented implemented Gpt3.5 and they did not use RAG. So the majority of Gpt3.5-focussed criticism so far cannot be extended to better performing LLMs

It is imperative to assess the strengths and limitations of using LLMs for evaluation and to consider comparative analyses to identify the most suitable model for specific educational needs. While LLMs are not yet perfect evaluators, they offer significant support in enhancing evaluation processes, including providing individualized feedback for students and teachers—a topic not covered in this article but deserving further exploration.

The findings of this article illustrate the vast potential of LLMs in educational evaluation and highlight the critical need for stakeholders in educational technology to engage in thorough research on the deployment of generative AI tools, or at least to be aware of the latest research results. Future research should explore also other LLMs, alternative prompting techniques, specialized training for specific evaluation tasks, and the application of these tools across various cultural and linguistic contexts. This comprehensive approach will ensure responsible and effective use of LLMs in educational environments, improving learning outcomes and supporting teachers in their educational tasks while upholding fairness and accuracy in evaluations regarding students' performance.